\documentclass[10pt,twocolumn,letterpaper]{article}

\usepackage{cvpr}
\usepackage{times}
\usepackage{epsfig}
\usepackage{graphicx}
\usepackage{amsmath}
\usepackage{amssymb}
\usepackage{booktabs} 
\usepackage{algorithm}
\usepackage{algorithmic}
\usepackage{subfigure}
\usepackage{url}
\usepackage{multirow}
\usepackage[usenames,dvipsnames,table]{xcolor}

\usepackage[pagebackref=true,breaklinks=true,letterpaper=true,colorlinks,bookmarks=false]{hyperref}

\cvprfinalcopy 

\def\cvprPaperID{98} 

\ifcvprfinal\fi
\begin{document}
	
	\title{Semantic Object Parsing with Local-Global Long Short-Term Memory}

	\author{Xiaodan~Liang$^{\dagger}$ $^{\star}$ \quad Xiaohui Shen$^{\ast}$ \quad Donglai Xiang$^{\dagger}$ \quad Jiashi Feng$^{\dagger}$ \quad Liang~Lin$^{\star}$ \quad Shuicheng~Yan $^{\dagger}$\\
		$^{\dagger}$ National University of Singapore \quad $^{\star}$ Sun Yat-sen University \quad $^{\ast}$ Adobe Research\\
		{\tt\small xdliang328@gmail.com  \quad xshen@adobe.com  \quad  xiangdonglai@u.nus.edu}\\
		{\tt\small elefjia@nus.edu.sg \quad linliang@ieee.org \quad eleyans@nus.edu.sg}
	}
	
	\maketitle
	
%
\begin{abstract}
   Semantic object parsing is a fundamental task for understanding objects in detail in computer vision community, where incorporating multi-level contextual information is critical for achieving such fine-grained pixel-level recognition. Prior methods often leverage the contextual information through post-processing predicted confidence maps. In this work, we propose a novel deep Local-Global Long Short-Term Memory (LG-LSTM) architecture to seamlessly incorporate short-distance and long-distance spatial dependencies into the feature learning over all pixel positions. In each LG-LSTM layer, local guidance from neighboring positions and global guidance from the whole image are imposed on each position to better exploit complex local and global contextual information. Individual LSTMs for distinct spatial dimensions are also utilized to intrinsically capture various spatial layouts of semantic parts in the images, yielding distinct hidden and memory cells of each position for each dimension. In our parsing approach, several LG-LSTM layers are stacked and appended to the intermediate convolutional layers to directly enhance visual features, allowing network parameters to be learned in an end-to-end way. The long chains of sequential computation by stacked LG-LSTM layers also enable each pixel to sense a much larger region for inference benefiting from the memorization of previous dependencies in all positions along all dimensions. Comprehensive evaluations on three public datasets well demonstrate the significant superiority of our LG-LSTM over other state-of-the-art methods. 
   
\end{abstract}


\section{Introduction}

\begin{figure}
	\begin{center}
		\includegraphics[scale=0.5]{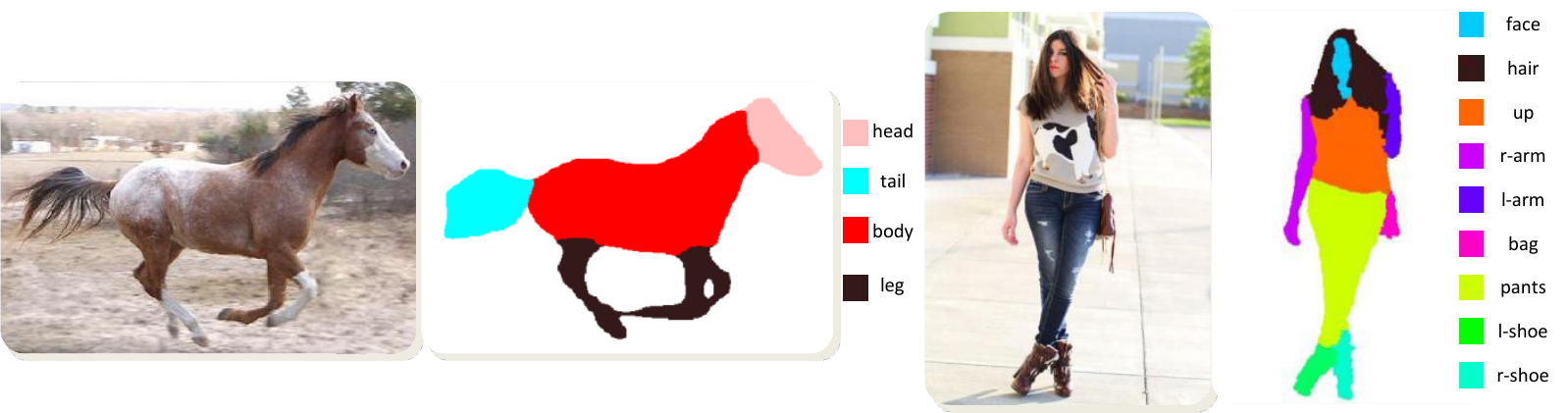}
		\caption{{Examples for semantic object parsing by our LG-LSTM architecture. Best viewed in color.}}
		\label{fig:task}
	\end{center}
	\vspace{-9mm}
\end{figure}

Semantic object parsing, which refers to segmenting an image region of an object into several semantic parts, enables the computer to understand the contents of an image in detail, as illustrated in Figure~\ref{fig:task}. It helps many higher-level computer vision applications, such as image-to-caption generation~\cite{chen2014learning}, clothes recognition and retrieval~\cite{Yamaguchiparsing13}, person re-identification~\cite{zhao2013unsupervised} and human behavior analysis~\cite{wang2012discriminative}.

Recently, many research works~\cite{ATR}\cite{wang2015joint}\cite{wang2014semantic}\cite{Co-CNN}\cite{ATR} have been devoted to exploring various Convolutional Neural Networks (CNN) based models for semantic object parsing, due to their excellent performance in image classification~\cite{szegedy2014going}, object segmentation~\cite{YannLepami13}\cite{papandreou2015weakly} and part localization~\cite{deeppose13}. However, the classification of each pixel position by CNNs can only leverage very local information from limited neighboring context, depicted by small convolutional filters. Intuitively, larger local context and global perspective of the whole image are very critical cues to recognize each semantic part in the image. For instance, in terms of local context, visually similar regions can be predicted as ``left-leg" or ``right-leg" depending on their specific locations and neighboring semantic parts (\eg ``left-shoes" or ``right-shoes"), especially for regions with two crossed legs. Similarly, the regions of ``tail" and ``leg" can be distinguished by the spatial layouts relative to the region of ``body". In terms of global perspective, distinguishing ``skirt" from ``dress" or ``pants" needs the guidance from the prediction on other semantic regions such as ``upper-clothes" or ``legs".  Previous works often resort to some post-processing techniques to separately address these complex contextual dependencies, such as, super-pixel smoothing~\cite{ATR}, mean field approximation~\cite{liu2015semantic} and conditional random field~\cite{chen2014semantic}\cite{crfasrnn}\cite{wang2015joint}. They improve accuracy through carefully designed processing on the predicted confidence maps instead of explicitly increasing the discriminative capability of visual features and networks. These separate steps often make feature learning inefficient and result in suboptimal prediction for pixel-wise object parsing.

Instead of employing separate processing steps, this work aims to explicitly increase the capabilities of features and networks in an end-to-end learning process. One key bottleneck to increase the network capability is the long-chain problem in deep CNN structures, that is, information from previous computations rapidly attenuates as it progresses through the chain. Similar problem exists when recurrent neural networks are applied to long sequential data. LSTM recurrent neural networks~\cite{lstm} were originally introduced to address this problem, which utilize the memory cells to process sequential data with complex and sequentialized interdependencies by independently reading, writing and forgetting some information. It could be similarly extended to image analysis. Particularly, the emergence of Grid LSTM~\cite{gridlstm} allows for multi-dimensional spatial communications. Our work builds on Grid LSTM~\cite{gridlstm} and proposes a novel Local-Global LSTM (LG-LSTM) for CNN-based semantic object parsing in order to simultaneously model global and local contextual information for improving the network capability. The proposed LG-LSTM layers are appended to the intermediate convolutional layers in a Fully Convolutional Neural Network~\cite{long2014fully} to enhance visual features by seamlessly incorporating long-distance and short-distance dependencies. The hidden cells in LG-LSTM serve as the enhanced features, and the memory cells serve as the intrinsic states that recurrently remember all previous interactions of all positions in each layer. 

To incorporate local guidance, in each LG-LSTM layer, the features at each position are influenced by the hidden cells at that position in the previous LG-LSTM layer (\ie depth dimension) as well as the hidden cells from eight neighboring positions (\ie spatial dimensions). The depth LSTM along the depth dimension is used to communicate information directly from one layer to the next while the spatial LSTMs along spatial dimensions allow for the complex spatial interactions and memorize previous contextual dependencies. Individual memory cells for each position are used for each of the dimensions to capture the diverse spatial layouts of semantic parts in different images.

Moreover, to further incorporate global guidance, the whole hidden cell maps obtained from the previous LG-LSTM layer are split into nine grids, with each grid covering one part of the whole image. Then the max-pooling over each grid selects discriminative features as global hidden cells, which are used to guide the prediction on each position. In this way, the global contextual information can thus be conveniently harnessed together with the local spatial dependencies from neighboring positions to improve the network capability.

By stacking multiple LG-LSTM layers and sequentially performing learning and inference, the prediction of each position is implicitly influenced by the continuously updated global contextual information about the image and the local contextual information of neighboring regions in an end-to-end way. During the training phase, to keep the network invariant to spatial transformations, all gate weights for the spatial LSTMs for local contextual interactions are shared across different positions in the same LG-LSTM layer, and the weights for the spatial LSTMs and the depth LSTM are also shared across different LG-LSTM layers. 

\begin{figure*}
	\begin{center}
		\includegraphics[scale=0.68]{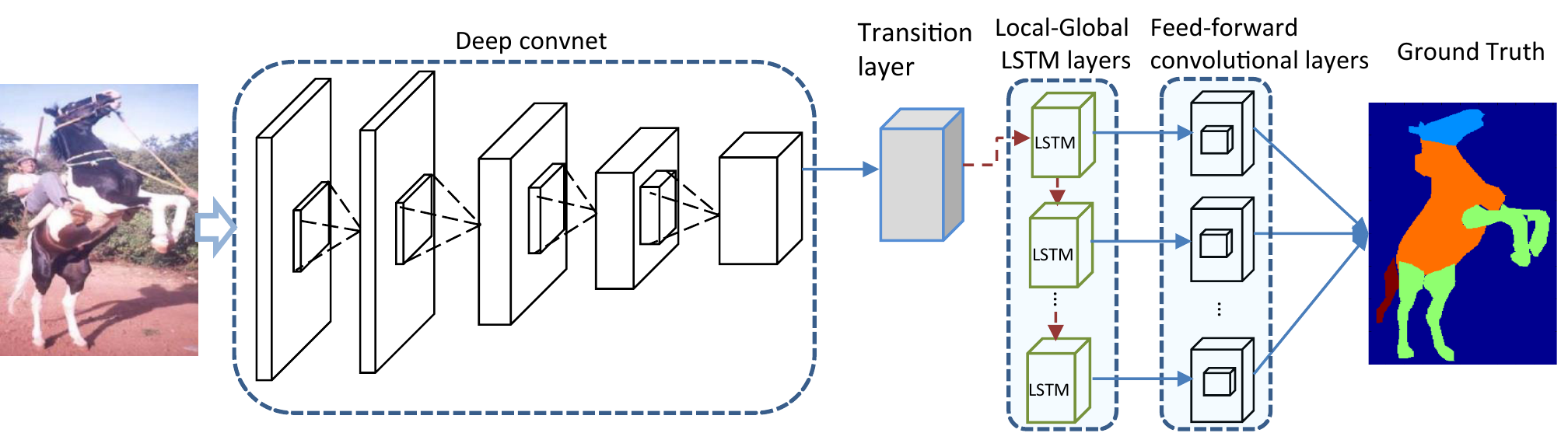}
		\vspace{-1mm}
		\caption{{The proposed LG-LSTM architecture. LG-LSTM integrates several novel local-global LSTM layers into the CNN architecture for semantic object parsing. An input image goes through several convolutional layers to generate its feature maps. Then the transition layer and several stacked LG-LSTM layers are appended to continuously improve the feature capability. Based on these enhanced feature maps, the feed-forward convolutional layer attached to the last LG-LSTM layer  produces the final object parsing result. The individual cross-entropy loss over all pixels is used to supervise the updating of each LG-LSTM layer. 
				}}
		\label{fig:framework}
	\end{center}
	\vspace{-6mm}
\end{figure*}


Our main contributions can be summarized in four aspects. 1) The proposed LG-LSTM exploits local contextual information to guide the feature learning of each position by using eight spatial LSTMs and one depth LSTM. 2) The global hidden cells are posed as the input states of each position to leverage long-distance spatial dependencies of the whole image. 3) The stacked LG-LSTM layers allow long-range contextual interactions benefiting from the memorization of previous dependencies in all positions. 4) The proposed LG-LSTM layers are incorporated into fully convolutional networks to enable an end-to-end feature learning over all positions. We conduct comprehensive evaluations and comparisons on the Horse-Cow parsing dataset~\cite{wang2014semantic}, and two human parsing datasets (\ie ATR dataset~\cite{ATR} and Fashionista dataset~\cite{yamaguchi2012parsing}). Experimental results demonstrate that our architecture significantly outperforms previously published methods for object parsing. 


\vspace{-2mm}
\section{Related Work}

\textbf{Semantic Object Parsing:} There have been increasing research works in the semantic object parsing problem including the general object parsing~\cite{wang2014semantic}\cite{wang2015joint}\cite{lu2014parsing}\cite{chen2014detect}\cite{hariharan2014hypercolumns} and human parsing~\cite{yamaguchi2012parsing}\cite{Yamaguchiparsing13}\cite{Dongparsing13}\cite{ICCV11WBW}\cite{SimoSerraACCV2014}\cite{M-CNN}. Recent state-of-the-art approaches often rely on deep convolutional neural networks along with advanced network architectures~\cite{hariharan2014hypercolumns}\cite{wang2015joint}\cite{Co-CNN}. 
Instead of learning features only from local convolutional
kernels as in these previous methods, we solve this problem through incorporating the novel LG-LSTM layers into CNNs to capture both long-distance and short-distance spatial dependencies. In addition, the adoption of hidden and memory cells in LSTMs makes it possible to memorize the previous contextual interactions from local neighboring positions and the whole image in previous LG-LSTM layers. It should be noted that while~\cite{crfasrnn} models mean-field approximate inference as recurrent networks, it can only refine results based on predicted pixel-wise confidence maps. In contrast, our LG-LSTM layers can progressively improve visual features to directly boost the performance of object parsing.

\textbf{LSTM on Image Processing:} LSTM networks have been successfully applied to many tasks such as hand-writing recognition~\cite{graves2009offline}, machine translation~\cite{sutskever2014sequence} and image-to-caption generation~\cite{showtell}. They have been further extended to multi-dimensional learning and applied to image processing tasks~\cite{byeon2014texture}~\cite{theis2015generative} such as biomedical image segmentation~\cite{stollenga2015parallel}, person detection~\cite{stewart2015end} and scene labeling~\cite{byeon2015scene}. Most recently, Grid LSTM~\cite{gridlstm} extended LSTM cells to allow the multi-dimensional communication across the LSTM cells, and the stacked LSTM and multi-dimensional LSTM~\cite{MDLSTM} can be regarded as special cases of Grid LSTM. The proposed LG-LSTM architecture in this work is extended from Grid LSTM and adapted to the complex semantic object parsing task. Instead of pure local factorized LSTMs in~\cite{gridlstm}~\cite{MDLSTM}, the features of each position in the proposed LG-LSTM are influenced by the short-distance dependencies as well as the long-distance global information from the whole image. Most of previous works verified the capability of Grid LSTM on very simple data (simple digital images, graphical or texture data), while we focus particularly on the higher-level object parsing task. The closest work to our method is the scene labeling approach proposed in~\cite{byeon2015scene}, where 2D LSTM cells are performed on the non-overlapping image patches. However, our architecture differs from that work in that we employ eight spatial LSTMs and one depth LSTM on each pixel, and learn distinct gate weights for different LSTMs by considering different spatial interdependencies. In addition, global hidden cells are also incorporated as the inputs for different LSTMs in each LG-LSTM layer. 

\vspace{-2mm}
\section{The Proposed LG-LSTM Architecture}

\subsection{Overview}

The proposed LG-LSTM aims to generate the pixel-wise semantic labeling for each image. As illustrated in Figure~\ref{fig:framework}, the input image is first passed through a stack of convolutional layers to generate a set of convolutional feature maps.  Then the transition layer adapts convolutional feature maps into the inputs of LG-LSTM layers, which are fed into the first LG-LSTM layer. The LG-LSTM layers is able to memorize long-period of the context information from local neighboring positions and global view of the image. More details about the LG-LSTM layers are presented in Section~\ref{sec:lstm}. After each LG-LSTM layer, one feed-forward convolutional layer  with $1\times1$ filters generates the $C$ confidence maps based on these improved features. The individual cross-entropy loss function over all pixels is used after each feed-forward convolutional layer in order to train each LG-LSTM layer. Finally, after the last LG-LSTM layer, the $C$ confidence maps for $C$ labels (including background) are inferred by the last feed-forward layer to produce the final object parsing result. 

\begin{figure*}
	\begin{center}
		\includegraphics[scale=0.65]{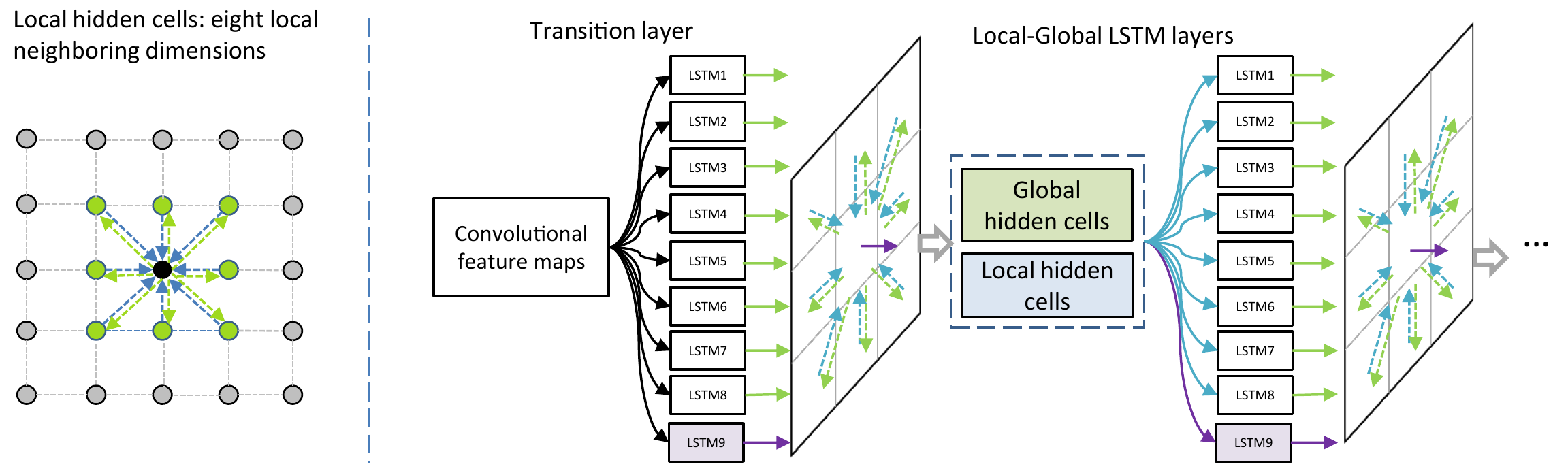}
		\caption{{Illustration of the proposed LG-LSTM layer. For local interactions, the prediction of each pixel (black point) depends on the features of all eight neighboring positions (green points), shown in the left column. The proposed LG-LSTM layer is shown in the right column. To connect the convolutional layers with LG-LSTM layers, we first feed convolutional feature maps into nine LSTMs to generate corresponding hidden cells. Then, local hidden cells comprise the hidden cells passed from eight neighboring directions (blue arrows) and that position (purple arrows) in the previous layer, or from the transition layer for the first LG-LSTM layer. Global hidden cells are constructed from hidden cells of all positions. We feed them both into nine LSTMs to produce individual hidden cells along different spatial dimensions (green arrows) and the depth dimension (purple arrow). Through recurrent connections by using stacked layers, the feature of each pixel can capture context information from a much larger local region and the global perspective of the whole image. Best viewed in color.}}
		\label{fig:lstm}
	\end{center}
	\vspace{-8mm}
\end{figure*}

\textbf{Transition Layer.} To make sure the number of the input states for the first LG-LSTM layer is consistent with that of the following LG-LSTM layers so that they can share all gate weights, the feature maps from convolutional layers are first adapted by the transition layer and then fed into the LG-LSTM layers. The transition layer uses the same number of LSTMs as in LG-LSTM layers, and passes the convolutional features of each position into these LSTMs to generate individual hidden cells. These resulting hidden cells are then used to construct input states for the first LG-LSTM layer. These weight matrices in the transition layer are not shared with those of the LG-LSTM layers because their dimensions of input states are not consistent. In the initialization, all memory cells are set as zeros for all positions, following the practical settings used in pedestrian detection~\cite{stewart2015end}. The updated memory cells from the transition layer can then be shared with LG-LSTM layers, which enables LG-LSTM layers to memorize feature representations obtained from convolutional layers. 

\subsection{Local-Global LSTM Layers}
\label{sec:lstm}

In this section, we describe the novel LG-LSTM layer tailored for semantic object parsing. To be self-contained, we first recall the standard LSTM recurrent neural network~\cite{lstm} and then describe the proposed LG-LSTM layers.

\vspace{-5mm}
\subsubsection{One-dimensional LSTM}

The LSTM network~\cite{lstm} easily memorizes the long-period interdependencies in sequential data. In image understanding, this temporal dependency learning can be conveniently converted to the spatial domain. The stacked layers of feature maps also enable the memorization of previous states at each pixel position (referred as depth dimension in this work). Each LSTM accepts the previous input $\mathbf{x}_i$ and determines the current states that comprises the hidden cells $\mathbf{h}_{i+1}\in \mathbb{R}^d$ and the memory cells $\mathbf{m}_{i+1} \in \mathbb{R}^d$, where $d$ is the output number. Following~\cite{graves2013speech}, the LSTM network consists of four gates: the input gate $g^u$, the forget gate $g^f$, the memory gate $g^c$ and the output gate $g^o$. The $W^u, W^f, W^c, W^o$ are the corresponding recurrent gate weight matrices. Suppose, $\mathbf{H}_i$ is the concatenation of the input $\mathbf{x}_i$ and the previous states $\mathbf{h}_{i}$. The hidden and memory cells can be updated as

\vspace{-6mm}
\begin{equation}
\begin{split}
g^u &= \delta(W^u * \mathbf{H}_i),\\
g^f &= \delta(W^f * \mathbf{H}_i),\\
g^o &= \delta(W^o * \mathbf{H}_i),\\
g^c &= \tanh(W^c * \mathbf{H}_i),\\
\mathbf{m}_{i+1} &= g^f \odot \mathbf{\mathbf{m}}_i + g^u \odot g^c,\\
\mathbf{h}_{i+1} &= \tanh(g^o \odot \mathbf{m}_i),
\end{split}
\label{eq:lstm}
\end{equation}
where $\delta$ is the logistic sigmoid function, and $\odot$ indicates a pointwise product. Let $\mathbf{W}$ denote the concatenation of four weight matrices. Following~\cite{gridlstm}, we use the function $\text{LSTM}(\cdot)$ to shorten Eqn.~(\ref{eq:lstm}) as

\vspace{-3mm}
\begin{equation}
(\mathbf{h}_{i+1}, \mathbf{m}_{i+1}) = \text{LSTM}(\mathbf{H}_i, \mathbf{m}_i, \mathbf{W}).
\end{equation}

The mechanism acts as a memory and implicit attention system, whereby the information from the previous inputs can be written to the memory cells and used to communicate with sequential inputs.

\begin{figure}
	\begin{center}
		\includegraphics[scale=0.82]{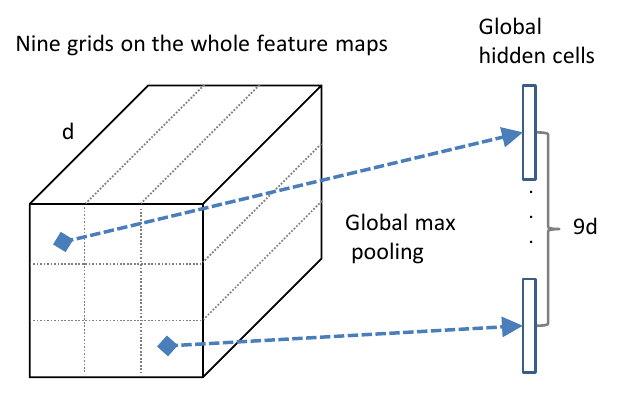}
		\vspace{-2mm}
		\caption{{Global hidden cells in LG-LSTM. The whole hidden cell maps are spatially partitioned into nine grids and the global max pooling is performed on each grid to generate the global hidden cells, which well capture global contextual information.}}
		\label{fig:global}
	\end{center}
	\vspace{-9mm}
\end{figure}

\vspace{-2mm}
\subsubsection{Local-Global LSTM Layers}

Grid LSTM~\cite{gridlstm} extended one-dimensional LSTM to cells that are arranged in a multi-dimension grid. Inspired by the design of Grid LSTM, we propose the multi-dimensional Local-Global LSTM to adapt to higher-level image processing. The features of each position depend on the local short-distance and global long-distance information. Local information propagated from neighboring pixels can help retain the short-distance contextual interactions (\eg object boundaries) while global information obtained from the whole feature maps can provide the long-distance contextual guidance (\eg global spatial layouts of semantic parts) to boost feature prediction of each position.

\textbf{Local Hidden Cells.} In terms of local interactions, as illustrated in Figure~\ref{fig:lstm}, the feature prediction of each position $j$ takes in $N=8$ hidden cells from $N$ local neighboring pixels from $N$ spatial LSTMs and the hidden cells from one depth LSTM. Intuitively, the depth LSTM can help track the previous information at each position using the memory cells benefited from the LSTM mechanism, like in the one-dimension LSTM. Each spatial LSTM computes the propagated hidden cells starting from each position to its corresponding spatial direction, as illustrated by the green arrows in Figure~\ref{fig:lstm}. Thus, each position can provide distinct guidance to each spatial direction by employing distinct spatial LSTMs, which take the spatial layouts and interactions into account for feature prediction. Let $\mathbf{{h}}^s_{i,j,n}\in \mathbb{R}^d, n \in \{1, \cdots, N\}$ denote the hidden cells propagated from the corresponding neighboring position to a specific pixel $j$ along the $n$-th spatial dimension by the $n$-th spatial LSTM, obtained from the $i$-th layer. The $\mathbf{{h}}^e_{i,j} \in \mathbb{R}^d$ indicates the hidden cell computed by the depth LSTM on the position $j$ using the weights updated in the $i$-th layer.

\textbf{Global Hidden Cells.} In terms of global interaction, the feature prediction of each position $j$ also takes the global hidden cells generated based on the whole hidden cell maps from the previous LG-LSTM layer as inputs. Specifically, the whole hidden cell maps are constructed by the hidden cells of the depth LSTM, i.e. $\mathbf{{h}}^e_{i,j}$ of all positions, which represents the enhanced features of each position. As shown in Figure~\ref{fig:global}, we partition the whole hidden cell maps into nine grids and then the global max-pooling within each grid is performed, resulting in $3\times 3$ hidden cell maps. The hidden cells for each grid can thus capture the global information of each grid. Such partition and max-pooling is perform over all the $d$ channels, forming $9\times d$ hidden cells. We denote the global hidden cells obtained from the $i$-th LG-LSTM layer as $\mathbf{f}_i \in \mathbb{R}^{9d}$. The global max pooling of hidden cells enables the model to seamlessly incorporate global and local information based on previous hidden cells in the previous LG-LSTM layer.

\textbf{LG-LSTM Layer.} Given the global hidden cells $\mathbf{f}_i$, the local hidden cells $\{\mathbf{{h}}^s_{i,j,n}\}_1^N$ from $N$ spatial LSTMs and $\mathbf{{h}}^e_{i,j}$ from one depth LSTM for each position $j$, the input states $\mathbf{H}_{i,j} \in \mathbb{R}^{(9+N+1)\times d}$ fed into the $(i+1)$-th LG-LSTM layer at each position $j$ can be computed as

\vspace{-4mm}

\begin{equation}
\mathbf{H}_{i,j} = [\mathbf{f}_i \ \  \mathbf{{h}}^s_{i,j,1} \ \  \mathbf{{h}}^s_{i,j,2} \  \dots \  \mathbf{{h}}^s_{i,j, N} \  \  \mathbf{{h}}^e_{i,j}]^T.
\label{eq:hiddeninput}
\end{equation}
\vspace{-3mm}

Denote memory cells of all $N$ spatial dimensions for each position $j$ as $\{\mathbf{m}^s_{i,j,n}\}_1^N \subset \mathbb{R}^d$ in the $i$-th LSTM layer and those of depth dimension as $\mathbf{m}^e_{i,j} \in \mathbb{R}^d$. Extended from Grid LSTM~\cite{gridlstm}, the new hidden cells and memory cells of each position $j$ for all $N+1$ dimensions are calculated as
  
\vspace{-6mm}
\begin{equation}
\begin{split}
(\hat{\mathbf{h}}^s_{i+1,j, 1}, \mathbf{m}^s_{i+1,j, 1}) &= \text{LSTM}(\mathbf{H}_{i,j}, \mathbf{m}^s_{i,j, 1}, \mathbf{W}^s_{i}),\\
(\mathbf{h}^s_{i+1,j, 2}, \mathbf{m}^s_{i+1,j, 2}) &= \text{LSTM}(\mathbf{H}_{i,j}, \mathbf{m}^s_{i,j, 2}, \mathbf{W}^s_{i}),\\
&\vdots\\
(\hat{\mathbf{h}}^s_{i+1,j, N}, \mathbf{m}^s_{i+1,j, N}) &= \text{LSTM}(\mathbf{H}_{i,j}, \mathbf{m}^s_{i,j, N}, \mathbf{W}^s_{i}),\\
({\mathbf{h}}^e_{i+1,j}, \mathbf{m}^e_{i+1,j}) &= \text{LSTM}(\mathbf{H}_{i,j}, \mathbf{m}^e_{i,j}, \mathbf{W}^e_{i}),
\end{split}
\vspace{-2mm}
\end{equation}
where $\hat{\mathbf{h}}^s_{i+1,j, n}$ represents the hidden cells propagated from the position $j$ to the $n$-th spatial direction, which are used by its neighboring positions to generate their input hidden cells in the next layer. Note that $\hat{\mathbf{h}}^s_{i+1,j, n}$ can be distinguished from the ${\mathbf{h}}^s_{i+1,j,n}$ by the different starting points and directions for information propagation. Each position thus has $N$ sides of incoming hidden cells and $N$ sides of outgoing hidden cells for incorporating complex local interactions. Although the same $\mathbf{H}_{i,j}$ for each position is applied for all LSTMs, the distinct memory cells for  each position used in different LSTMs enable individual information propagation from $N$-sides. These LSTM functions are operated on all positions to produce the whole hidden cell maps. To keep invariance along different spatial dimensions, the weight matrices $\mathbf{W}^s_{i}$ of $N$ spatial LSTMs are shared. By sequentially stacking several LSTM layers, the receptive field of each position can be considerably increased to sense a much larger contextual region. In addition, long-distance information can also be effectively captured by using the global hidden cells. The input features fed into feed-forward convolutional layers are computed as $\mathbf{H}_{i,j}$ for each position.


\vspace{-2mm}
\section{Experiments}

\subsection{Experimental Settings}

\textbf{Dataset:} We evaluate the performance of LG-LSTM architecture for semantic object parsing on the Horse-Cow parsing dataset~\cite{wang2014semantic} and two human parsing datasets, ATR dataset~\cite{Co-CNN} and Fashionista dataset~\cite{yamaguchi2012parsing}. 

\textbf{Horse-Cow parsing dataset~\cite{wang2014semantic}.} The Horse-Cow parsing dataset is a part segmentation benchmark introduced in~\cite{wang2014semantic}. For each class, most observable instances from PASCAL VOC 2010 benchmark~\cite{everingham2012pascal} are manually selected, including 294 training images and 227 testing images. Each image pixel is elaborately labeled as one of the four part classes, including head, leg, tail and body. Following the experiment protocols in~\cite{wang2014semantic} and~\cite{wang2015joint}, we use all the training images to learn the model and test every image given the object class. The standard intersection over union (IOU) criterion and pixel-wise accuracy are adopted for evaluation. We compare the results of our LG-LSTM with three state-of-the-art methods, including the compositional-based method (``SPS")~\cite{wang2014semantic}, the hypercolumn (``HC")~\cite{hariharan2014hypercolumns} and the most recent method (``Joint")~\cite{wang2015joint}.

\textbf{ATR dataset~\cite{ATR} and Fashionista dataset~\cite{yamaguchi2012parsing}.} Human parsing aims to predict every pixel of each image with 18 labels: face, sunglass, hat, scarf, hair, upper-clothes, left-arm, right-arm, belt, pants, left-leg, right-leg, skirt, left-shoe, right-shoe, bag, dress and null. Originally, 7,700 images are included in the ATR dataset~\cite{ATR}, with 6,000 for training, 1,000 for testing and 700 for validation. 10,000 real-world human pictures are further collected by~\cite{Co-CNN} to cover images with more challenging poses, occlusion and clothes variations. We follow the training and testing settings used in~\cite{Co-CNN}. The Fashionista dataset contains 685 images, among which 229 images are used for testing and the rest for training. We use the same evaluation metrics as in~\cite{Yamaguchiparsing13}~\cite{ATR}~\cite{Co-CNN}, including accuracy, average precision, average recall, and average F-1 score. We compare the results of our LG-LSTM with five recent state-of-the-art approaches~\cite{yamaguchi2012parsing}~\cite{Yamaguchiparsing13}~\cite{M-CNN}~\cite{ATR}~\cite{Co-CNN}.


\begin{table}\setlength{\tabcolsep}{3.8pt}
	\centering\scriptsize
	\caption{Comparison of object parsing performance with three state-of-the-art methods over the Horse-Cow object parsing dataset~\cite{wang2014semantic}. We report the IoU accuracies on background class, each part class and foreground class. }\label{tab:horsecow}
	\begin{tabular}{cccccccccccccccccccccc}
		\toprule
		& & & & \textbf{Horse} & & &\\
		\hline
		{Method} &  Bkg   &  head  &  body  & leg & tail & Fg & IOU & Pix.Acc \\
		\midrule
		SPS~\cite{wang2014semantic}  & 79.14 & 47.64 & 69.74 & 38.85 & - & 68.63 & - & 81.45 \\
		HC~\cite{hariharan2014hypercolumns}  & 85.71 & 57.30 & 77.88 & 51.93 & 37.10 & 78.84 & 61.98 & 87.18 \\
		Joint~\cite{wang2015joint} & 87.34 & 60.02 & 77.52 & 58.35 & \textbf{51.88} & 80.70 & 65.02 & 88.49\\
		\hline
		\textbf{LG-LSTM} & \textbf{89.64} & \textbf{66.89} & \textbf{84.20} & \textbf{60.88} & 42.06 & \textbf{82.50} & \textbf{68.73} & \textbf{90.92} \\
		\midrule
		& & & & \textbf{Cow} & & &\\
		\hline
		{Method} &  Bkg   &  head  &  body  & leg & tail & Fg & IOU & Pix.Acc \\
		\midrule
		SPS~\cite{wang2014semantic}  & 78.00 & 40.55 & 61.65 & 36.32 & - & 71.98 & - & 78.97 \\
		HC~\cite{hariharan2014hypercolumns}  & 81.86 & 55.18 & 72.75 & 42.03 & 11.04 & 77.04 & 52.57 & 84.43 \\
		Joint~\cite{wang2015joint} & 85.68 & 58.04 & 76.04 & 51.12 & 15.00 & 82.63 & 57.18 & 87.00\\
		\hline
		\textbf{LG-LSTM} & \textbf{89.71} & \textbf{68.43} & \textbf{82.47} & \textbf{53.93} & \textbf{19.41} & \textbf{85.41} & \textbf{62.79} & \textbf{90.43}\\
		\hline
	\end{tabular}%
	\vspace{-6mm}
\end{table}%

\textbf{Implementation Details:} In our experiments, five LG-LSTM layers are appended to the convolutional layers right before the prediction layer of the basic CNN architecture. For fair comparison with~\cite{wang2015joint}, we fine-tune the network based on the publicly available pre-trained VGG-16 classification network for the Horse-Cow parsing dataset. We utilize the slightly modified ``DeepLab-CRF-LargeFOV" network structure  presented in~\cite{chen2014semantic} as the basic architecture due to its leading accuracy and competitive efficiency. This network architecture~\cite{chen2014semantic} transforms VGG-16 ImageNet model to fully-convolutional network and changes the number of filters in the last two layers from 4096 to 1024. For evaluation on two human parsing datasets, the basic ``Co-CNN" structure proposed in~\cite{Co-CNN} is utilized due to its leading accuracy. In terms of training based on ``Co-CNN", our model is trained from the scratch following the same training and testing settings in~\cite{Co-CNN}. Our code is implemented based on the publicly available Caffe platform~\cite{jia2014caffe} and all networks are trained on a single NVIDIA GeForce GTX TITAN X GPU with 12GB memory. 

We use the same settings for data augmentation techniques for the object part segmentation and human parsing as in~\cite{wang2015joint} and~\cite{Co-CNN}, respectively. The scale of input image is fixed as $321\times321$ for training models based on VGG-16 for object part segmentation. During training based on ``Co-CNN", the input image is rescaled to $150\times 100$ following the same settings in~\cite{Co-CNN}. During fine-tuning, the learning rate of the newly added layers, including transition layer, LG-LSTM layers and feed-forward convolutional layers is initialized as 0.001 and that of other previously learned layers is initialized as 0.0001. For training based on ``Co-CNN", the learning rate of all layers is initialized as 0.001. All weight matrices used in the LG-LSTM layers are randomly initialized from a uniform distribution of [-0.1, 0.1]. In each LG-LSTM layer, eight spatial LSTMs for different spatial dimensions and one depth LSTM are deployed in each position, and each LSTM predicts $d = 64$ dimension hidden cells and memory cells. We only use five LG-LSTM layers for all models since significant improvements are not observed by using more LG-LSTM layers which also cost more computation resources. The weights of all convolutional layers are initialized with Gaussian distribution with standard deviation as 0.001. We train all the models using stochastic gradient descent with a batch size of 2 images, momentum of 0.9, and weight decay of 0.0005. We fine-tune the networks on VGG-16 for roughly 60 epochs and it takes about 1 day for the Horse-Cow parsing dataset. For training based on ``Co-CNN" from the scratch, it takes about 4-5 days for the human parsing datasets. In the testing stage,  one image takes 0.3 second on average.

\vspace{-2mm}
\subsection{Results and Comparisons}
\vspace{-1mm}

We compare the proposed LSTM architecture with the strong baselines on three public datasets.

\begin{table}\setlength{\tabcolsep}{2pt}
	\centering\scriptsize
	\caption{Comparison of human parsing performance with five state-of-the-art methods when evaluating on ATR~\cite{ATR}.}\label{tab:tableoverall}
	\begin{tabular}{cccccccccccccccccccccc}
		\toprule
		\textbf{Method} &  \textbf{Acc.}   &  \textbf{F.g. acc.}  &  \textbf{Avg. prec.}   &    \textbf{Avg. recall}  &  \textbf{Avg. F-1 score} \\
		\midrule
		Yamaguchi et al.~\cite{yamaguchi2012parsing}& 84.38 & 55.59 & 37.54 & 51.05 & 41.80 \\
		PaperDoll~\cite{Yamaguchiparsing13} & 88.96 & 62.18 & 52.75 & 49.43 & 44.76 \\
		{M-CNN}~\cite{M-CNN} &{89.57} &{73.98} &{64.56} &{65.17} &{62.81}\\
		ATR~\cite{ATR} & {91.11} & {71.04} & {71.69} & {60.25} & {64.38}\\
		{Co-CNN}~\cite{Co-CNN} & 95.23 & 80.90 & 81.55 & 74.42 & 76.95\\
		{Co-CNN (more)}~\cite{Co-CNN} & {96.02} & {83.57} & {84.95} & {77.66} & {80.14}\\
		\midrule
		{LG-LSTM} & {96.18} & {84.79} & {84.64} & {79.43} & {80.97}\\
		\textbf{LG-LSTM (more)} & \textbf{96.85} & \textbf{87.35} & \textbf{85.94} & \textbf{82.79} & \textbf{84.12}\\
		\bottomrule
	\end{tabular}%
		\vspace{-4mm}
\end{table}%

\begin{table}\setlength{\tabcolsep}{2pt}
	\centering\scriptsize
	\caption{Comparison of human parsing performance with four state-of-the-art methods on the test images of Fashionista~\cite{yamaguchi2012parsing}.}\label{tab:tablefashion}.
	\begin{tabular}{cccccccccccccccccccccc}
		\toprule
		\textbf{Method} &  \textbf{Acc.}   &  \textbf{F.g. acc.}  &  \textbf{Avg. prec.}   &    \textbf{Avg. recall}  &  \textbf{Avg. F-1 score} \\
		\midrule
		Yamaguchi et al.~\cite{yamaguchi2012parsing}  & 87.87 & 58.85 & 51.04 & 48.05 & 42.87 \\
		PaperDoll~\cite{Yamaguchiparsing13}  & 89.98 & 65.66 & 54.87 & 51.16 & 46.80 \\
		ATR~\cite{ATR} & {92.33} & {76.54} & {73.93} & {66.49} & {69.30}\\
		{Co-CNN}~\cite{Co-CNN} & 96.08 & 84.71 & 82.98 & 77.78 & 79.37\\
		{Co-CNN (more)}~\cite{Co-CNN} & {97.06} & {89.15} & {87.83} & {81.73} & {83.78}\\
		\midrule
		{LG-LSTM} & {96.85} & {87.71} & {87.05} & {82.14} & {83.67}\\
		\textbf{LG-LSTM (more)} & \textbf{97.66} & \textbf{91.35} & \textbf{89.54} & \textbf{85.54} & \textbf{86.94}\\
		\bottomrule
	\end{tabular}%
	\vspace{-6mm}
\end{table}%

\textbf{Horse-Cow Parsing dataset~\cite{wang2014semantic}}: Table~\ref{tab:horsecow} shows the performance of our models and comparisons with three state-of-the-art methods on the overall metrics. The proposed LG-LSTM architecture can outperform three baselines with significant gains: $9.47\%$ over SPS~\cite{wang2014semantic}, $3.74\%$ over HC~\cite{hariharan2014hypercolumns}, $2.43\%$ over Joint~\cite{wang2015joint} in terms of overall pixel accuracy for the horse class. Our method also gives a huge boost in average IOU: LG-LSTM achieves $68.73\%$, $6.75\%$ better than HC~\cite{hariharan2014hypercolumns} and $3.71\%$ better than Joint~\cite{wang2015joint} for the horse class. The large improvement, i.e. $5.61\%$ increase by LG-LSTM in IOU over the best performing state-of-the-art method, can also be observed from the comparisons on cow class. This superior performance achieved by LG-LSTM demonstrates that utilizing stacked LG-LSTM layers is very effective in capturing the complex contextual patterns within images that are critical for distinguishing and segmenting different semantic parts from an instance with homogeneous appearance such as a horse or a cow. 

\textbf{ATR dataset~\cite{ATR}}: Table~\ref{tab:tableoverall} and Table~\ref{tab:F1scores} show the performance of our models and comparisons with five state-of-the-arts on overall metrics and F-1 scores of individual semantic labels, respectively. The proposed LG-LSTM can significantly outperform five baselines, particularly, $80.97\%$ vs $64.38\%$ of ATR~\cite{ATR} and $76.95\%$ of Co-CNN~\cite{Co-CNN} in terms of average F-1 score. Following~\cite{Co-CNN}, we also take the additional 10,000 images in~\cite{Co-CNN} as the supplementary training images and report the results as ``LG-LSTM (more)". The ``LG-LSTM (more)" can also improve the average F-1 score by $3.98\%$ over ``Co-CNN (more)". We show the F-1 scores for each label in Table~\ref{tab:F1scores}. Generally, our LG-LSTM shows much higher performance than other methods. In terms of predicting semantic labels with small regions such as hat, belt, bag and scarf, our method offers a very large gain. This demonstrates that our LG-LSTM architecture  performs very well on this human parsing task. 

\begin{figure}
	\begin{center}
		\includegraphics[scale=0.5]{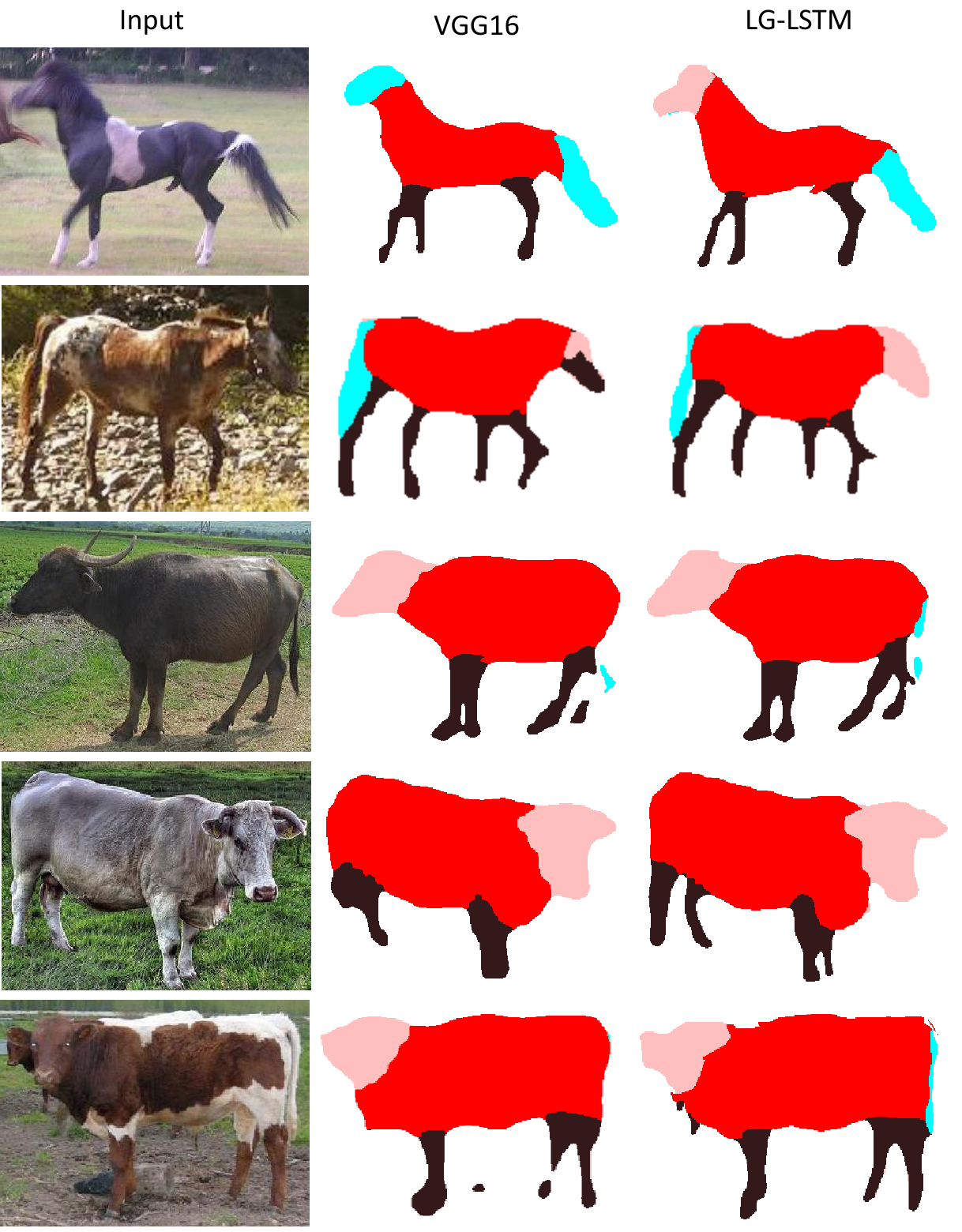}
		\caption{{Comparison of object parsing results by our LG-LSTM architecture and the baseline that only fine-tunes the model based on VGG-16 network. Best viewed in color.}}
		\vspace{-6mm}
		\label{fig:parsingresults}
	\end{center}
\end{figure}

\textbf{Fashionista dataset~\cite{yamaguchi2012parsing}}: Table~\ref{tab:tablefashion} gives the comparison results on the 229 test images of the Fashionista dataset. All results of the state-of-the-art methods were reported in~\cite{ATR} and~\cite{Co-CNN}. Following~\cite{ATR}, we only report the performance by training on the same large ATR dataset~\cite{ATR} and then testing on the 229 images of the Fashionista dataset. Our LG-LSTM architecture can substantially outperform the baselines by the large gains over all metrics.

\begin{table*}\setlength{\tabcolsep}{4pt}
	\centering\scriptsize
	{\caption{Per-Class Comparison of F-1 scores with five state-of-the-art methods on  ATR~\cite{ATR}.}\label{tab:F1scores}
		
		\begin{tabular}{cccccccccccccccccccccc}
			\toprule
			Method & Hat & Hair & S-gls & U-cloth & Skirt & Pants & Dress & Belt & L-shoe  & R-shoe & Face & L-leg & R-leg & L-arm  & R-arm & Bag & Scarf   \\
			\midrule
			Yamaguchi et al.~\cite{yamaguchi2012parsing} & 8.44 & 59.96 & 12.09 & 56.07 & 17.57 & 55.42 & 40.94 & 14.68 & 38.24 & 38.33 & 72.10 & 58.52 & 57.03 & 45.33 & 46.65 & 24.53 & 11.43\\
			PaperDoll~\cite{Yamaguchiparsing13}& 1.72 & 63.58 & 0.23 & 71.87 & 40.20 & 69.35 & 59.49 & 16.94 & 45.79 & 44.47 & 61.63 & 52.19 & 55.60 & 45.23 & 46.75 & 30.52 & 2.95\\
		    {M-CNN}~\cite{M-CNN}  & {80.77} & 65.31 & 35.55& {72.58}& 77.86 &	70.71 &	81.44&	38.45& 	53.87&{48.57}&	72.78&	63.25&{68.24}&{57.40}& 	{51.12}&{57.87}& 43.38\\
			{ATR~\cite{ATR}} & {77.97} & 68.18 & {29.20} & 79.39 & {80.36} & {79.77} & \textbf{82.02} & 22.88 & {53.51} & {50.26} & 74.71 & {69.07} & {71.69} & {53.79} & {58.57} & 53.66 & \textbf{57.07}\\
			Co-CNN~\cite{Co-CNN} & 72.07 & 86.33 & 72.81 & 85.72 & 70.82 & 83.05 & 69.95 & 37.66 & 76.48 & 76.80 & 89.02 & 85.49 & 85.23 & 84.16 & 84.04 & 81.51 & 44.94\\
			{Co-CNN more}~\cite{Co-CNN} & {75.88} & {89.97} & \textbf{81.26} & {87.38} & {71.94} & {84.89} & {71.03} & 40.14 & {81.43} & {81.49} & {92.73} & {88.77} & {88.48} & {89.00} & {88.71} & {83.81} & {46.24}\\
			\midrule
			\textbf{LG-LSTM (more)}	& \textbf{81.13} & \textbf{90.94} & {81.07} & \textbf{88.97} & \textbf{80.91} & \textbf{91.47} & {77.18} & \textbf{60.32} & \textbf{83.40} & \textbf{83.65} & \textbf{93.67} & \textbf{92.27} & \textbf{92.41} & \textbf{90.20} & \textbf{90.13} & \textbf{85.78} & 51.09\\			
			\bottomrule
		\end{tabular}
		\vspace{-4mm}
	}
	
\end{table*}

	\begin{figure*}
		\begin{center}
			\includegraphics[scale=0.86]{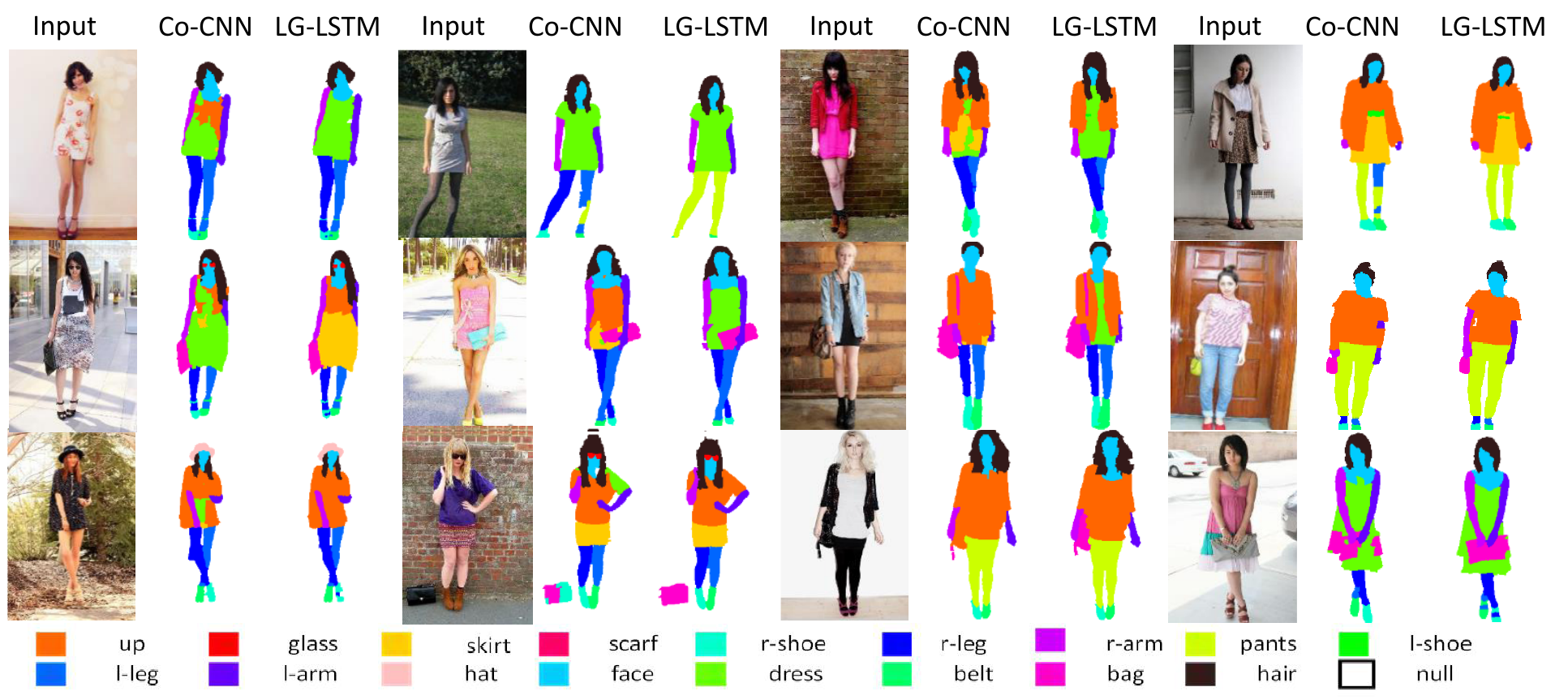}\vspace{-2mm}
			\caption{{Comparison of object parsing results of our LG-LSTM architecture and the Co-CNN~\cite{Co-CNN} on ATR dataset~\cite{ATR}. }}
			\vspace{-8mm}
			\label{fig:humanparsing}
		\end{center}
	\end{figure*}
	
	\begin{table}\setlength{\tabcolsep}{2pt}
		\centering\scriptsize
		\caption{Comparison of object parsing performance with different variants of the proposed LG-LSTM architecture on Horse-Cow object parsing dataset~\cite{wang2014semantic}.}\label{tab:ablation}
		\begin{tabular}{cccccccccccccccccccccc}
			\toprule
			& & & & \textbf{Horse} & & &\\
			\hline
			{Method} &  Bkg   &  head  &  body  & leg & tail & Fg & IOU & Pix.Acc \\
			\midrule
			VGG16 & 87.73 & 60.25 & 80.06 & 56.74 & 35.87 & 80.19 & 64.13 & 89.00\\
			VGG16 extra conv  & 88.60 & 62.02 & 81.39 & 58.10 & 39.63 & 80.92 & 65.95 & 89.80\\
			\hline
			LG-LSTM local\_2 & 88.58 & 64.63 & 82.53 & 58.53 & 39.33 & 81.32 & 66.72 & 89.99\\ 
			LG-LSTM local\_4 & 88.97 & 66.43 & 83.42 & 59.49 & 40.06 & 81.64 & 67.68 & 90.41\\ 
			LG-LSTM w/o global & 89.06 & 65.18 & 83.07 & 59.26 & 40.76 & 81.71 & 67.46 & 90.39\\
			\hline
			\textbf{LG-LSTM} & \textbf{89.64} & \textbf{66.89} & \textbf{84.20} & \textbf{60.88} & 42.06 & \textbf{82.50} & \textbf{68.73} & \textbf{90.92} \\
			\midrule
			& & & & \textbf{Cow} & & &\\
			\hline
			{Method} &  Bkg   &  head  &  body  & leg & tail & Fg & IOU & Pix.Acc \\
			\midrule
			VGG16 & 87.62 & 57.97 & 75.39 & 47.21 & 13.89 & 81.78 & 56.41 & 87.73\\
			VGG16 extra conv  & 87.46 & 60.88 & 76.60 & 47.82 & 16.93 & 82.69 & 57.93 & 87.83\\
			\hline
			LG-LSTM local\_2 & 87.57 & 61.06 & 76.82 & 48.32 & 19.21 & 82.78 & 58.60 & 87.95\\ 
			LG-LSTM local\_4 & 88.14 & 63.34 & 78.57 & 49.90 & 19.29 & 83.54 & 59.85 & 88.67\\ 
			LG-LSTM w/o global & 88.77 & 64.59 & 80.11 & 52.21 & 19.24 & 84.19 & 60.98 & 89.41\\
			\hline
			\textbf{LG-LSTM} & \textbf{89.71} & \textbf{68.43} & \textbf{82.47} & \textbf{53.93} & \textbf{19.41} & \textbf{85.41} & \textbf{62.79} & \textbf{90.43}\\
			\bottomrule
		\end{tabular}%
		\vspace{-6mm}
	\end{table}%
	
\vspace{-1mm}
\subsection{Discussions}
\vspace{-1mm}

We further evaluate different network settings to verify the effectiveness of the important components in our LG-LSTM architecture, presented in Table~\ref{tab:ablation}.

\textbf{Comparison with using convolutional layers}: To strictly evaluate the effectiveness of using the proposed LG-LSTM layers, we report the performance of purely using convolutional layers, i.e., ``VGG16", indicating the results of the basic network architecture we use with one extra feed-forward convolution layer with $1\times 1$ filters attached to output pixel-wise confidence maps. By comparing ``VGG16" with ``LG-LSTM", $4.6\%$ improvement in IOU on horse class can be observed, which demonstrates the superiority of using more LG-LSTM layers to jointly address the long-distance and short-distance contextual information. Note that, since the LSTMs are deployed in each position in each LG-LSTM layer and more parameters are introduced for model learning, one possible alternative solution is just using more convolutional layers instead of LG-LSTM layers. We thus report the performance of using more convolutional layers on the basic network structure, i.e. ``VGG16 extra conv". To make fair comparison with our usage of five LG-LSTM layers, five extra convolutional layers are utilized containing $576 = 64 \times 9$ convolutional filters with size $3\times3$ in each convolutional layer, because nine LSTMs are used in LG-LSTM layers and each of them has 64 hidden cell outputs. Compared with ``LG-LSTM", the ``VGG16 extra conv" decreases the mean IOU by $2.78\%$ and 
$4.86\%$ on horse and cow classes, respectively. It speaks well for the superiority of using LG-LSTM layers to harness complex long-distances patterns and memorize long-period hidden states over purely convolutional layers. 

\textbf{Local connections in LG-LSTM}: Note that in LG-LSTM, we use eight spatial neighboring connections to capture local contextual information for each position. To further validate the effectiveness of LG-LSTM, we also report the performance of using two local connections and four local connections, i.e. ``LG-LSTM local\_2" and ``LG-LSTM local\_4".  For ``LG-LSTM local\_2", the top and left neighboring positions with respect to each position are used. For ``LG-LSTM local\_4", the local information is propagated from four neighboring positions in top, left, top-left and top-right directions. It can be observed that our LG-LSTM architecture (``LG-LSTM") significantly outperforms ``LG-LSTM local\_2" and ``LG-LSTM local\_4" by $4.19\%$ and $2.94\%$ in IOU on cow class, respectively. Eight spatial connections with neighboring pixels for each position enable LG-LSTM to capture richer information from neighboring context, which are more informative than other limited local spatial connections. As illustrated in the first row of Figure~\ref{fig:humanparsing}, LG-LSTM gives more consistent parsing results by incorporating sufficient local connections while Co-CNN produces many fragments of semantic regions. 

\textbf{Global connections in LG-LSTM}: ``LG-LSTM w/o global" in Table~\ref{tab:ablation} indicates the results without using global hidden cells as the LSTM inputs for each position. Compared with ``LG-LSTM", $1.27\%$ and $1.81\%$ decreases in IOU occur with ``LG-LSTM w/o global" on horse and cow classes, respectively. This demonstrates well the effectiveness of global contextual information for inferring the prediction of each position. These global features provide an overview perspective of the image to guide the pixel-wise labeling. As shown in the second row of Figure~\ref{fig:humanparsing}, by gathering global information from the whole image, LG-LSTM successfully distinguishes the combination of upper-clothes and skirt with dress while Co-CNN often confuses them only from local cues. 

\vspace{-2mm}
\subsection{More Visual Comparison}

The qualitative comparisons of parsing results on Horse-Cow dataset and ATR dataset are visualized in Figure~\ref{fig:parsingresults} and Figure~\ref{fig:humanparsing}, respectively. Because the previous state-of-the-art methods do not publish their codes, we only compare our method with the ``VGG16" on Horse-Cow parsing dataset. As can be observed from these visualized comparisons, our LG-LSTM architecture outputs more semantically meaningful and precise predictions than ``VGG16" and ``Co-CNN" despite the existence of large appearance and position variations. For example, the small regions (\eg tails) can be successfully segmented out by LG-LSTM from neighboring similar semantic regions (\ie body and legs) on the Horse-Cow dataset. LG-LSTM can successfully handle the confusing labels such as skirt vs dress and legs vs pants on the human parsing dataset. The regions with similar appearances can be recognized and separated by the guidance from global contextual information, while the local boundaries for different semantic regions are preserved well by using local connections.

\vspace{-2mm}
\section{Conclusions and Future Work}
In this work, we proposed a novel local-global LSTM architecture for semantic object parsing. The LG-LSTM layers jointly capture the long-distance and short-distance spatial dependencies by using global hidden cells from the whole maps and local hidden cells from eight spatial dimensions and one depth dimension. Extensive results on three public datasets clearly demonstrated the effectiveness of the proposed LG-LSTM in generating pixel-wise semantic labeling. In the future, we will explore how to develop a pure LG-LSTM network architecture where all convolutional layers are replaced with well designed LG-LSTM layers. It will produce more complex neural units in each layer, which can hierarchically exploit local and global connections of the whole image, and enables to remember long-period hidden states to capture complex visual patterns.
					
\vspace{-3mm}
{\small
\bibliographystyle{ieee}
\bibliography{egbib}
}

\end{document}